\documentclass[wcp]{jmlr}

\usepackage{booktabs}
\usepackage{graphicx}
\usepackage{threeparttable}
\usepackage{times}
\usepackage{amsmath}
\usepackage{soul}
\usepackage{url}
\usepackage[utf8]{inputenc}
\usepackage{amsfonts}
\usepackage{graphicx}
\usepackage{amsmath, bm}
\usepackage{booktabs}
\usepackage{algorithm}
\usepackage{algorithmic}
\usepackage{url}

\newcommand{\Comment}[1]{\hfill $\vartriangleright$ {#1}}
\newcommand{\BComment}[1]{\Comment{\textcolor{blue}{#1}}}
\newcommand{\RComment}[1]{\Comment{\textcolor{red}{#1}}}
\newcommand{\GComment}[1]{\Comment{\textcolor{green}{#1}}}

\newcommand{\tabincell}[2]{\begin{tabular}{@{}#1@{}}#2\end{tabular}}

\newcommand{\hide}[1]{}
\newcommand{\atn}[1]{\textcolor{black}{#1}}
\newcommand{\atnn}[1]{\textcolor{black}{#1}}

\newcommand{\reminder}[1]{{\textsf{\textcolor{blue}{[#1]}}}} 
\newcommand{\mkclean}{\renewcommand{\reminder}{\hide}}
\mkclean

\usepackage{xspace}
\newcommand{\method}{MissGAN\xspace}

\newtheorem{iproblem}{Informal Problem}

\newlength\myindent
\makeatletter
\let\Ginclude@graphics\@org@Ginclude@graphics 
\makeatother

\jmlrvolume{}
\jmlryear{2021}
\jmlrworkshop{ACML 2021}

\title[]{Multi-scale Anomaly Detection for Big Time Series of Industrial Sensors}

 \author{\Name{Quan Ding} \Email{dingquan20@mails.ucas.ac.cn}\\
  \Name{Shenghua Liu} \Email{liushenghua@ict.ac.cn}\\
  \Name{Bin Zhou} \Email{zhoubin17g@ict.ac.cn}\\
  \Name{Huawei Shen} \Email{shenhuawei@ict.ac.cn}\\
  \Name{Xueqi Cheng} \Email{cxq@ict.ac.cn}\\
  \addr Institute of Computing Technology, Chinese Academy of Sciences}

\begin{document}

\maketitle

\begin{abstract}
    Given a multivariate big time series, can we detect anomalies
    as soon as they occur? Many existing works detect anomalies
    by learning how much a time series deviates away from what it
    should be in the reconstruction framework. However, most models
    have to cut the big time series into small pieces empirically
    since optimization algorithms cannot afford such a long series.
    The question is raised: \emph{do such cuts pollute the inherent
        semantic segments, like incorrect punctuation in sentences?}

    Therefore, we propose a reconstruction-based anomaly detection
    method, \method, iteratively learning to decode and encode
    naturally smooth time series in coarse segments, and finding
    out a finer segment from low-dimensional representations based on
    HMM. As a result, learning from multi-scale segments, \method can
    reconstruct a meaningful and robust time series, with the help of
    adversarial regularization and extra conditional states. \method
    does not need labels or only need labels of normal instances,
    making it widely applicable. Experiments on industrial datasets of
    real water network sensors show our \method outperforms the baselines
    with scalability. Besides, we use a case study on the CMU Motion dataset
    to demonstrate that our model can well distinguish unexpected gestures
    from a given conditional motion.
\end{abstract}
\begin{keywords}
    Big Time Series; Reconstruction; Anomaly Detection;
    Multi-scale Training; Segmentation; GAN Network; GRU
\end{keywords}

\section{Introduction}


Big time series are generated from countless domains, such as
infrastructure, system monitoring, personal wearable devices,
and medical analysis~\cite{faloutsos2019forecasting}.
\atnn{While big time series always have a long length,}
detecting the anomalies in such multivariate time series is a key
to secure infrastructures and systems functioning,
and diagnose the anomalies of people's motion and health data.
However, traditional supervised machine learning methods cannot
handle this task properly, because of
the inherent nature that labeled anomalies are far
fewer, and anomalies arise differently from each other, i.e.,~
obtaining accurate and representative features \atnn{is} challenging
~\cite{chandola2009anomaly}.
Thus the question is raised:

\emph{
	How can we  detect anomalies in big time series, when observing
	only normal time series or observed data being normal most of the time?
}


Some existing works~\cite{kiat2018doping,wang2019wgan} augment anomalous
instances from labeled anomalies to balance the training. Those
methods assume that the existing labeled anomalies are representative,
without considering unseen types of anomalies.
The non-supervised methods, either learning only from labeled
normal instances or only from unlabeled data (containing very few
anomalies) perform promisingly and are widely applicable
in anomaly detection~\cite{chalapathy2019deep,chandola2009anomaly,hooi2017b}.
Among those methods, the reconstruction of data is the most frequently
used framework, and anomalies produce high reconstruction error~\cite{shah2014spotting}.
Auto-encoders (AE)~\cite{han2011data} allows for more complex patterns
by applying nonlinear functions for reconstruction and anomaly detection.
Moreover, combined with GAN~\cite{goodfellow2014generative}, the performance of the
encoder-decoder model improves further via adversary regularization.

GAN is used widely on recontruction-based anomaly detection task.
AnoGAN~\cite{schlegl2017unsupervised} is the first
application of GAN on medical images whose running consumes a great deal
of time. Later work Ganomaly~\cite{akcay2018ganomaly} and
EGBAD~\cite{zenati2018efficient} focus on adding a coding part for an end-to-end
model. In terms of time series anomaly detection, GAN-based models~\cite{zhou2019beatgan,li2019mad}
reconstructed the given time series segments
for anomaly detection, for example, real-valued medical time
series~\cite{esteban2017real}. 
Variant of computing the loss of GAN is applied in
BeatGAN~\cite{zhou2019beatgan} which performs well on ECG data. MAD-GAN~\cite{li2019mad}
combines LSTM-RNN with the GAN framework and reports good results on the SWaT dataset.
\atnn{However, its inefficiency in calculating the best match for each test case limits
	its application. 
	}
\atnn{Most of these models use the sliding window algorithm to segment big time series
	which may produce pathologically poor results under
	some circumstances~\cite{keogh2004segmenting}.} Thus, the second question is raised:

\emph{
	How can we find out a group of cutting points that follows the inherent characteristics
	of big time series data?
}

\atnn{Multi-scale segmentation and feature extraction are broadly used in image processing~\cite{568922}.
	\cite{zeune2017multiscale} uses multi-scale segmentation on
	images to find multiple objects with different scales.
	Transferring the idea into time series, \cite{cho2012multiscale} tries to locate breakpoints in different scales.}
	AutoPlait~\cite{matsubara2014autoplait} and its variant Neucast~\cite{chen2018neucast} use
	the HMM-based model and MDL principle to make segmentations.

Therefore, we propose \method, simultaneously \underline{m}ult\underline{i}-\underline{s}cale reconstruction and
\underline{s}egmentation for big time series anomaly detection (see Fig~\ref{fig:architecture}).
Our method exploits extra conditional information to reconstruct multi-mode time series,
and outputs explainable results by reconstruction error, pinpointing the specific anomalous time ticks.
Experiments on time series from SWaT testbed and CMU Motion Capture data
show the effectiveness and robustness of our \method.

In summary, our main contributions are summarized as follows:
\atn{
	\begin{itemize}
		\item \textbf{Multi-scale reconstruction:} \method iteratively
		      learns to reconstruct from initially coarse and long
		      segments of time series,
		      and with learned hidden representation, MissGAN finds proper cuts on
		      current segments in turn to optimize reconstruction.
		      In such a way, reconstruction is gradually improved
		      by training on multi-scale segments of big time series, i.e.,~from coarse to
		      fine-grained.
		      Moreover, with conditional reconstruction,
		      \method can generate multi-mode time series given different states.
		\item \textbf{Effectiveness:}
		      Experiments on the publicly available data show that our method outperforms
		      the baselines, including both linear and non-linear
		      models in anomaly detection.
		      On the motion dataset, \method can be trained to
		      reconstruct well from the given walking and running time series
		      and discriminates against other types of unexpected gestures.
		\item \textbf{Explainability:}
		      \method can pinpoint the exact time ticks when anomalies occur in
		      a segment of time series, routing people's attention to diagnosis.
		\item \textbf{Scalability:}
		      Our method can detect anomalies in 1.78 ms/tick on average,
		      linear in the size of the total time series.
	\end{itemize}
}

\method is open-sourced for
reproducibility\footnote{ {https://www.dropbox.com/sh/pnn4mjbpltdlzbf/AADu9Brpym3WrHDwbrfLUEM4a?dl=0} }.






\section{Related Work}
The main purpose of anomaly detection is to identify anomalous cases that
deviate far from the distribution learned during the training with normal data.
Given the reality that labeled
anomaly data lacks, unsupervised algorithms are preferred.  Refer to~\cite{li2019mad}, 
anomaly detection algorithms can be classified into three
categories: i) linear model based method, ii) distance and probabilistic based
method and iii) deep learning based method.

\begin{table}[t]
    \caption{Comparison with related methods.}
    \centering
    \label{tab:related_work}
    \begin{tabular}{l|cccc}
        \hline
        methods                           & non-linear & explainability & extra conditions & \tabincell{c}{multi-scale \\segmentation} \\
        \hline
        {PCA~\cite{li2014model}}          &            & \checkmark     &                  &                          \\
        {KNN~\cite{angiulli2002fast}}     & \checkmark & \checkmark     &                  &                          \\
        {BeatGAN~\cite{zhou2019beatgan}}  & \checkmark & \checkmark     &                  &                          \\
        {LSTM-AE~\cite{malhotra2016lstm}} & \checkmark & \checkmark     &                  &                          \\
        {MAD-GAN~\cite{li2019mad}}        & \checkmark & \checkmark     & ?                &                          \\
        \hline
        {\method}                         & \checkmark & \checkmark     & \checkmark       & \checkmark               \\
        \hline
    \end{tabular}
\end{table}

\textbf{Linear methods.} Principal Component Analysis (PCA)~\cite{li2014model}
is the most familiar approach to most of us. As a multivariate data analysis
method, PCA extracts information and reduce dimensions from highly correlated
data by orthogonal transformation.

\textbf{Distance and probabilistic based methods.} K-Nearest Neighbor (KNN) is a popular method which
calculates the anomaly score by computing average distance to K nearest
neighbors~\cite{angiulli2002fast}. Although this method seems simple and
effective, we still need some prior knowledge to improve its performance, such
as numbers of anomalies or numbers of clusters. Yeh, $et$ $al.$ proposed a
parameter-free, fast and general algorithm Matrix Profile~\cite{yeh2016matrix}
to solve various time series problems. Another problem of distance based method
is how to segment time series properly. Probabilistic based method can be
regarded as upgrades of distance based methods with regarding to the data
distributions. For example, Feature Bagging (FB) 
method~\cite{lazarevic2005feature} pays attention to the correlations of
variables and performs well. Other works such as Hidden Markov
Models~\cite{baum1966statistical} is of great use for segmentation. ~\cite{molina2001clause} proposed methods to detect clause. Variations like
DynaMMo~\cite{li2009dynammo} and AutoPlait~\cite{matsubara2014autoplait} segmented
series on vocabulary-learning rules. Recent work like BeatLex~\cite{hooi2017b}
utilized Minimum Description Length (MDL) to learn vocabularies. These methods
have made progress compared to traditional sliding window methods. Yet,
distributions of temporal data are volatile and hard to observe and thus these
methods are not welcome in some applications.

\textbf{Deep Learning based methods} have made great improvements and gains so much
popularity ever since the boosting development of big data and deep learning
architectures. Autoencoder~\cite{han2011data} is used widely benefiting from its
ability of coding and reconstructing to catch features. LSTM-AE~\cite{malhotra2016lstm}
detects anomalies by reconstructing and calculating anomalousness score based
on LSTM cells.
Kieu $et$ $al.$~\cite{kieu2019outlier} propose ensemble frameworks based on
sparsely-connected RNN to do unsupervised outlier detection. \atnn{Xu $et$ $al.$
    propose Donut\cite{xu2018unsupervised}, which is also an autoencoder-based model
    designed for time series anomaly detection.}
Recently, the generative adversarial network has shown great ability in
learning data features and distributions. Therefore, it has been deployed on
image processing tasks, such as generating synthetic images~\cite{di2019survey}.
AnoGAN~\cite{schlegl2017unsupervised} is the first
application of GAN on medical images whose running consumes a great deal
of time. Later work Ganomaly~\cite{akcay2018ganomaly} and
EGBAD~\cite{zenati2018efficient} focus on adding a coding part for an end-to-end
model. Furthermore, more and more works pay attention to the application of GAN on
generating time series sequences, for example, real-valued medical time
series~\cite{esteban2017real}. Luo $et$ $al.$ propose $E^2$GAN~\cite{luo20192} to do
time series imputation. Reconstruction based anomaly detection method is applied in
BeatGAN~\cite{zhou2019beatgan} which performs well on ECG data. MAD-GAN~\cite{li2019mad}
combines LSTM-RNN with the GAN framework and reports good results on the SWaT dataset.
\atnn{However, its inefficiency in calculating the best match for each test case limits
    its application. Besides, Hundman $et$ $al.$ propose an unsupervised anomaly detection
    approach Telemanom\cite{hundman2018detecting} which uses LSTMs to predict highvolume
    telemetry data.}
Nevertheless, the aforementioned methods can only run on fixed-length segments and
cannot utilize conditional information.


Table~\ref{tab:related_work} summarizes the comparison of the related works
with our \method in the four characteristics. \atnn{We use a non-linear method to handle
    the more sophisticated dataset.
    Explainability requires results of methods can direct people's attention to anomalies.
    Extra conditions stand for the ability of the model to utilize extra
    information, i.e., labels.
    Multi-scale segmentation means whether the model can segment data dynamically. }
The question mark means that MADGAN
concatenates those extra conditions as input time series.
We can see that only \method meets all the characteristics.

\section{Proposed Model}
\begin{figure*}[t]
    \centering
    \includegraphics[width=1.0\textwidth, height=0.5\textwidth]{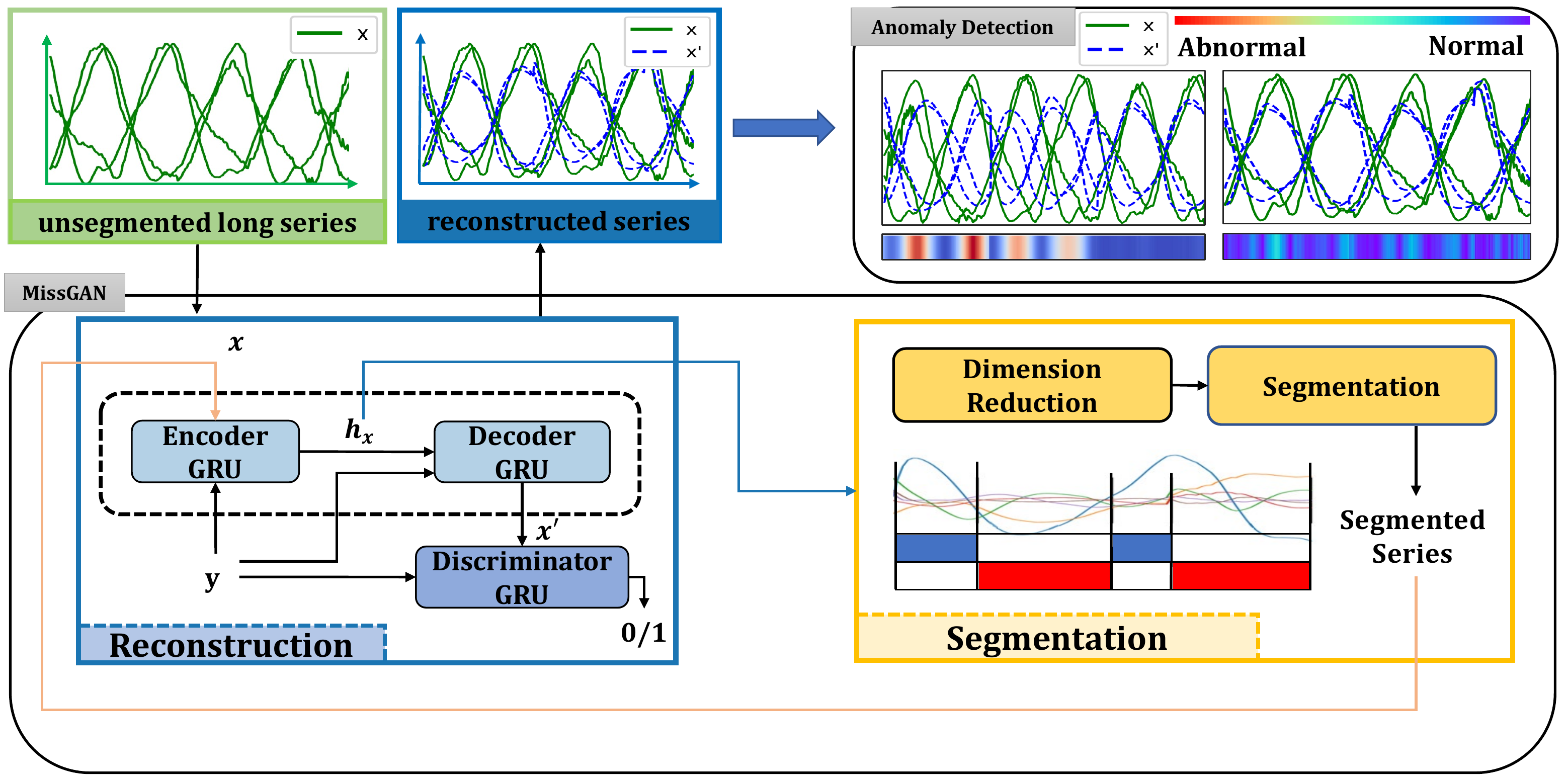}
    \caption{Overview of \method.}
    \label{fig:architecture}
\end{figure*}

Let \atn{$\bm {x} = \{x_0, x_1, \cdots\}$ be a multivariate time series,
    where each point ${x_i} \in \mathbf R^M$ } consists of $M$
dimensions which read from $M$ different sources at time $t_i$. A segment $\bm {x_j}$
is defined as
small fragment data extracted from $\bm {x}$ and denotes
as ${\bm {x}}_{\rho_j}^{\rho_j+l_j} \in
    \mathbf{R}^{M \times l_j}$ where $\rho_j$ is the start point and $l_j$ is the
length of the segment. Inside each segment $\bm {x_j}$ exists $M$
readings that record real-time data. We use $\bm {y}$ to stand for the categorical
data that is auxiliary to realize features and distributions.

The existing deep learning method shall always divide input series into fixed-length
segments which we believe may cause bad effects in training. As a result, our first
mission is to segment input series with a proper length $l$ to construct a collection of
segments $\mathcal S$. With segments divided properly, we can then finish our second
mission which is described as:

\atn{
    \begin{iproblem}[Anomalous time series detection]
        Given a big time series $\bm x$ of $M$ dimensions from daily
        monitoring of running systems or personal wearable sensors, and their
        states $\bm y$, knowing that
        at most of the time the systems or people are normal under states $\bm y$,
        \begin{itemize}
            \item \textbf{to} detect anomalies happening in time series $\bm x$,
            \item \textbf{such that} the anomalous time ticks of time series deviate far from
                  what they are supposed to be (reconstructed).
        \end{itemize}
    \end{iproblem}
}

\subsection{Framework Overview}
As Fig~\ref{fig:architecture} shows, our proposed model consists of two parts.
The first part is called reconstruction, which is responsible for training a network that
combines the discriminator of classic conditional GAN with an extra encoder-decoder
network being its reconstruction framework to minimize the reconstruction error as Eq~\eqref{equ:obj} shows. 
Details of the reconstruction model is introduced in Sec~\ref{sec:rec}. Furthermore, to explore an 
appropriate segmentation for a better reconstruction, we exploit an HMM-based segmenting algorithm 
which is introduced in Sec~\ref{sec:seg}.

\begin{equation}
    \begin{aligned}
    L = \left \| x - G_D(G_E(x)) \right\|_2
    \end{aligned}
    \label{equ:obj}
\end{equation}

In testing phase, to judge whether a segment $\bm{x_j}$ is anomalous, we reconstruct the segment $\bm{x_j}$ using our generator $G$
and calculate the anomalousness score. Because our model is trained by normal
data, we can assert that the segment deviates far from the normal distribution shall get a bad reconstruction, 
i.e., a relevant high anomalousness score shown in Eq~\eqref{equ:ano_score}, where $x_{j_t}$ is the data vector of time tick $t$
and $x_{j_t}'$ is the reconstructed data vector.

\begin{equation}
    \begin{aligned}
        \setlength{\belowdisplayskip}{3pt}
        \label{equ:ano_score}
        A(x_{j_t})=\left\| {x_{j_t}} - {x'_{j_t}} \right\|_2, x_{j_t} \in \bm{x_j}
    \end{aligned}
\end{equation}



\subsection{Reconstruction Model}
\label{sec:rec}

As illustrated in Fig~\ref{fig:architecture}, our reconstruction network consists of an encoder-decoder
framework and a discriminator of the classic GAN network. Both encoder and decoder are implemented by GRU.
Extra information, i.e., conditional dimension $\bm y$ is directly added to both the encoder and decoder
to take control of the reconstruction of different modes of data. So the total input for each GRU cell is
the concatenation of $x_t$ and $y_t$. The encoder $G_E(\bm{x})$ encodes the input $\bm x$ to a hidden
representation $\bm h$ to extract its features
 . The last hidden state of the encoder
is fed into the decoder as its first hidden state. And the decoder $G_D(\bm{x})$ reconstructs the time series $\bm {x'}$
in a reversed order.



The classical conditional GAN framework includes two parts:
the generative part $G$ is responsible for capturing the distribution of input data $p_{\bm{x}}$ and
the discriminative part is designed for estimating the probability that the input sample is concocted
by the generator rather than extracted from real data $\bm{x}$. In practice, we don't directly use
the classic loss function of the generator due to different frameworks of the generator.
Instead, we use pairwise feature matching loss designed for minimizing
the distance from the origin data to the generated time series.
Regard $f_D(\cdot)$ as the activation vector located at the hidden layer of the discriminator,
we combine the actual distance from origin time series $\bm{x}$ to
reconstructed time series $G_D(G_E(\bm{x}))$ with the pairwise feature matching loss
accompanied by a regularization parameter $\lambda$:


\begin{equation}
    \label{equ:gen_loss}
    \begin{aligned}
        L_G = \left \|\bm{x}-G_D(G_E(\bm{x})) \right\|_2 + \lambda \left \| f_D(\bm x|\bm y)-f_D(G_D(G_E(\bm{x}))|\bm y) \right\|_2
    \end{aligned}
\end{equation}

Meanwhile, the target of the discriminator $D$ is to reduce the probability that mistaking
reconstructed samples as origin samples. That's to maximize:

\begin{equation}
    \label{equ:dis_loss}
    \begin{aligned}
        L_D =\log D(\bm{x}|\bm{y})+\log (1-D({G_D(G_E(\bm{x}))}|\bm{y}))
    \end{aligned}
\end{equation}

\subsection{Segmentation Model}
\label{sec:seg}

We use a two-tier HMM-based method to find a set of cut points $p = \{\rho_1, \rho_2, \rho_3, ...\}$ for segmentation,
\atn{where the $regime$ is defined as a group of segments, and each segment has an assignment to one of the several regimes.}
Let $\theta$ be HMM model parameters for a regime, including initial state probability,
state transition probability, and output probability.
Regimes are then modeled by HMM with parameters, i.e., ${\theta_1, \cdots, \theta_r}$, and
regime transition matrix denotes as $\Delta_{r\times r}$, where $r$ is the regime size.



Model parameters are learned based on the MDL (minimum description length) principle to minimize the
total cost $Cost(\bm x, \Theta_H)$ shown in Eq~\ref{equ:autoplait}. This cost includes three parts:
$Cost_{model}$ describes the coding length of model parameters, $Cost_{assign}$ calculates the coding
length of pattern assignment and the cut points, and $Cost_{like}$ refers to the likelihood of such
assignment by a negative log-likelihood function. Besides, the construction of regimes plays a vital role 
in the segmentation task, a large granularity may concatenate several patterns into one regime and
a small granularity may produce several fractured regimes. So referring to~\cite{chen2018neucast}, 
we adapt the formula of calculating total 
cost by adding a hyper-parameter $\alpha$ for controlling the granularity of distinct patterns and assign a
default value of $0.1$. 
\begin{equation}
    \begin{aligned}
        \label{equ:autoplait}
        Cost(\bm{x};\Theta_H)=\alpha \times Cost_{model}(\Theta_H) +Cost_{assign}+Cost_{like}(\bm{x}|\Theta_H)
    \end{aligned}
\end{equation}

In general, we firstly preprocess origin data $\bm x$ and divide them coarsely into long series $\bm x_{init}$
with length manually assigned, i.e., $l_{init}$ and construct the collection of segments $\mathcal{S}$.
This initial length is always large enough to contain several periods of data, and we feed these segments
$\bm x_{init}$ into the reconstruction framework and fetch the latent-space representation $\bm {h_x}$
coded by its encoder part. Then, considering repetitive information may hide in the latent-space representation,
we reduce the dimension of hidden representation from $d_h$ to $d_r$ by PCA. The HMM-based segmentation model
will process the results to search for proper cut points making up of collection $p$. Finally,
we re-segment origin time series with the known cut points and feed back the newly segmented series
collection $\mathcal{S'}$ into reconstruction part and continue training to get a new updated latent-space
representation $\bm {h_x}$. With adequate iterations, we can extract the cut point data $p$ 
from the assigned result.
The final collection of segments $\mathcal{S}$ will then
be used to train the reconstruction network.

\subsection{Proposed {\method}}

\method first trains with coarse-segmented time series and outputs hidden representations as well as learns currently best
segmentation. In turn, these optimized segments are fed back to train reconstruction.
In such a way, the whole process is optimized until no more segmentation.

\begin{algorithm}[tb]
    \caption{\method Algorithm}
    \label{alg:overall}
    \begin{algorithmic}[1] 
        \STATE $\Theta_G, \Theta_D, \Theta_H {\leftarrow}$ initialize parameters
        \STATE Sample segments $\mathcal S(0) = \{ \bm{x_1},...,\bm{x_m} \} $ $\sim$ a batch of long fragments of time series
        \FOR{$k = 1, 2, \cdots, K$}
        \STATE $S(k) = S(k-1)$  \RComment{Segmentation Iterations}
        \FOR{$i = 1, 2, \cdots$}
        \STATE Sample $\{\bm{x_1}, \bm{x_2}, \cdots, \bm{x_j}\}$ from $\mathcal S(k)$   \RComment{Reconstruction Iterations}
        \STATE Reconstruct $\{\bm{x'_1}, \bm{x'_2}, \cdots, \bm{x'_j}\}$ by $G_E, G_D$ and $D$ \GComment{Reconstruction}
        \STATE Compute $L_D$ by Eq~\eqref{equ:dis_loss}
        \STATE $\Theta_D {\ \longleftarrow\ } \Theta_D +\beta \nabla_{\Theta_D}(L_D)$ \BComment{ $\nabla$ is the gradient}
        \STATE Compute $L_G$ by Eq~\eqref{equ:gen_loss}
        \STATE $\Theta_G {\ \longleftarrow\ } \Theta_G +\beta \nabla_{\Theta_G}(L_G)$ \BComment{ $\nabla$ is the gradient}
        \ENDFOR
        \STATE $\bm{h_x} = $ hidden representation in $G_E$
        \STATE $\mathcal S(k) = $ fine segments by minimizing Eq~\eqref{equ:autoplait} \GComment{Segmentation}
        \IF{${\mathcal S(k)} \equiv {\mathcal S(k-1)}$}
        \STATE \textbf{break}
        \ENDIF
        \ENDFOR
        \STATE $\bm{x'} = $ final training reconstruction model with last $\mathcal S$ as step 6-11
    \end{algorithmic}
\end{algorithm}

Let ${\mathcal S(k)} = \{\bm{x_1}, \bm{x_2}, \cdots\}$ be segmentation results in the $k$-th iteration.
Therefore, the overall reconstruction optimizes loss on multi-scale segments of time series, as follows.

\begin{equation}
    \nonumber
    \begin{aligned}
        L_G = \sum_{k=1}^{K}\sum_{\bm x \in \mathcal S(k)}^{}(
         & \left \|\bm{x}-G_D(G_E(\bm{x})) \right\|_2 +
         & \lambda \left \| f_D(\bm x|\bm y)-f_D(G_D(G_E(\bm{x}))|\bm y) \right\|_2
        )
    \end{aligned}
\end{equation}

\begin{equation}
    \nonumber
    \begin{aligned}
        L_D = \frac{1}{K}\sum_{k=1}^{K}\sum_{\bm x \in \mathcal S(k)}^{}[\log D(\bm x|\bm y)+\log (1-D(G_D(G_E(\bm x))|\bm y))]
    \end{aligned}
\end{equation}

Finally, A summary of the overall algorithm is depicted in Alg~\ref{alg:overall}.

\section{Experiment}
Experiments are designed to answer the following questions:

\noindent\textbf{Q1. Accuracy:} How accurately does our method find out
anomalies compared with baselines.

\noindent\textbf{Q2. Effectiveness and explainability:} How effectively does
\method find out anomalies in the data of both real-world system sensors and
personal wearable motion sensors?
How well does \method pinpoint anomalous time ticks of input time series
and route people’s attention?

\noindent\textbf{Q3. Scalability:} How fast does \method test on samples? What is the relation between running time and test sample length?

\noindent\textbf{Q4. Robustness and parameter sensitivity:} How sensitive does \method react to changes in parameters?

\begin{table}[tbp]
    \caption{General Description of Dataset.}
    \centering
    \begin{threeparttable}
        \label{tab:swat}
        \begin{tabular}{lcc}
            \hline
            Item                       & SWaT      & Motion  \\
            \hline
            Data Dimensions            & $25$      & $4$     \\
            Conditional Dimensions     & $26$      & $2$     \\
            Training Size (time ticks) & $496,800$ & $8,224$ \\
            Testing Size (time ticks)  & $449,919$ & $2,085$ \\
            Normal Rate\tnote{1}       & $88.02$   & $79.77$ \\
            \hline
        \end{tabular}
        \begin{tablenotes}
            \item[1] Normal Rate is the percentage of normal data in testing data
        \end{tablenotes}
    \end{threeparttable}
\end{table}

\subsection{Dataset}

We evaluate our proposed method on two datasets. The first one is the secure water treatment system (SWaT) dataset~\cite{mathur2016swat}.
A total of 25 dimensions that record readings of sensors are regarded as input dimensions while the other 26 dimensions which record states
of actuators are regarded as additional information, i.e., the conditional dimensions. The second dataset comes from a motion dataset captured by CMU.
This dataset includes motions such as walking, jumping, running, hopping, etc. recorded by 4 sensors, i.e., left and right arms and legs.
\atnn{As there are exact labels for each segment of running and walking, we regard the labels as conditional dimensions.}
Detailed information of the aforementioned datasets is depicted in Table~\ref{tab:swat}.

\subsection{Baselines and metrics}

The baselines include 
BeatGAN~\cite{zhou2019beatgan}, LSTM-AE~\cite{malhotra2016lstm} and MADGAN~\cite{li2019mad}. Parameters of these methods are adjusted well
to get their best performances.
Besides, we also implement CRGAN, which is \method without multi-scale segmentation, and AEGAN, which is \method without PCA processing
to do ablation experiments.

\method calculates the anomalousness score for each time tick in the evaluation dataset.
To make a comparison with baselines, we first standardize the anomalousness
score by min-max scaling to $0 \sim 1$. Then we use two metrics, AUC (Area Under ROC Curve) and ideal F1 score.
Given different thresholds, we get different precision and recall values. The best value will be treated as our ideal F1 score.


\begin{table}[htbp]
    \centering
    \caption{Performance of each method on AUC score and ideal F1 score based on the results of SWaT dataset, repeated 5 times.}
    \begin{threeparttable}

        \centering
        \label{tab:result_contrast}
        \begin{tabular}{lcc}
            \hline
            Method            &  AUC Score                   &Ideal F1 Score                \\
            \hline
            BeatGAN           &  $0.8143 \pm 0.0027$         &$0.7699 \pm 0.0109$           \\
            LSTM-AE           &  $0.8137 \pm 0.0077$         &$0.7780 \pm 0.0037$           \\
            MAD-GAN           &  $-$                         &$0.77$                        \\
            \hline
            CRGAN             &  $0.8217 \pm 0.0120$         &$0.7752 \pm 0.0034$           \\
            AEGAN             &  $0.8242 \pm 0.0120$         &$0.7830 \pm 0.0120$           \\
            \method           &  $\mathbf{0.8426 \pm 0.0060}$&$\mathbf{0.7844 \pm 0.0019}$  \\
            \method$_{0.5\%}$ &  $0.8381 \pm 0.0084$         &$0.7808 \pm 0.0023$           \\
            \method$_{1.0\%}$ &  $0.8348 \pm 0.0089$&$0.7799 \pm 0.0009$          
            \\
            \hline
        \end{tabular}
    \end{threeparttable}
\end{table}
\subsection{Accuracy and comparison (Q1)}

\paragraph{Experimental setup}

We choose GRU~\cite{chung2014empirical} with a single layer of 100 hidden neurons in the encoder, decoder, and discriminator structure.
Adam optimizer is used with the learning rate $\beta$ initialized as $0.001$, and decayed by 25\% for every 8 epochs.
We set the regularization parameter, $\lambda$ as $0.1$ according to results of parameter sensitive experiments.
We reduce the dimensions by PCA
from $d_h = 100$ to $d_r = 6$ before feeding to the segmentation model.
Granularity controlling hyper-parameter $\alpha$ in the segmentation model is set as 0.1 referred to~\cite{chen2018neucast}.

\paragraph{Results.} \atnn{Table~\ref{tab:result_contrast} shows the ideal F1 score and AUC score of \method and baselines.
    Results of MADGAN is extracted from~\cite{li2019mad}.
    \method outperforms all baseline methods on the ideal F1 score. About the AUC score, \method exceeds other
    baselines for at most $0.0289$.}

    \begin{figure*}[htbp]
        \centering
        \includegraphics[width=0.8\textwidth]{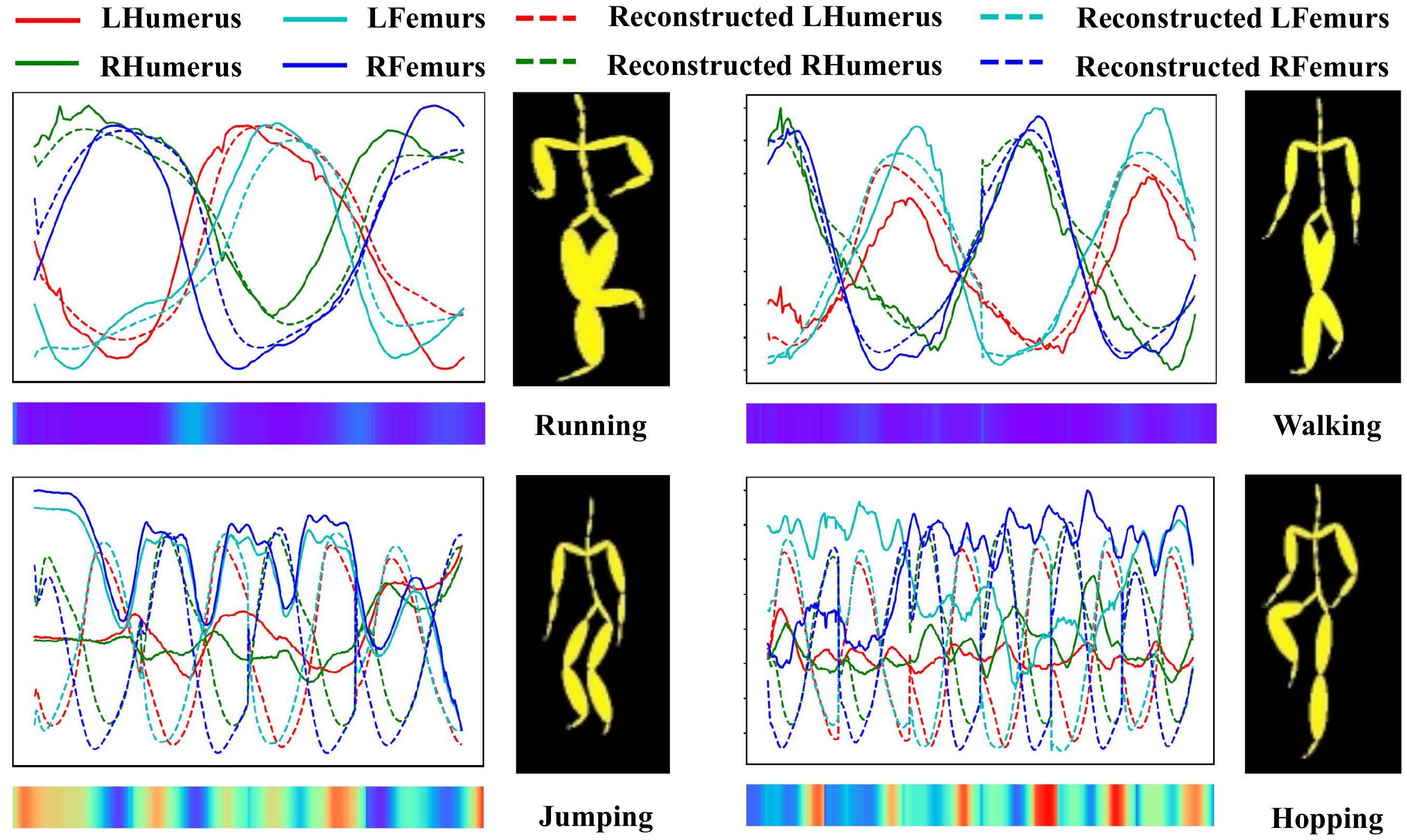}
        \label{fig:mocap_reconstruction}
        \caption{Reconstruction results of motion time series.}
    \end{figure*}
\atnn{CRGAN is \method without the segmentation part which is intended to show the effectiveness of segmentation.
    From Table~\ref{tab:result_contrast}, our proposed \method outperforms CRGAN both on ideal F1 score and AUC score which demonstrates
    multi-scale segmentation do make contributions to train the model.} AEGAN is \method with hidden dimensions in GRU equalling to the reduced dimension
after PCA processing in \method which demonstrates the effectiveness of dimension reduction by PCA. We also use this dataset to design experiments on
evaluating the robustness of our \method by adding anomalous cases ($0.5 \%$ and $1.0 \%$ of total time tick) to training data.

\subsection{Effectiveness and explainability (Q2)}

\atnn{We use Mocap dataset to do a case study to demonstrate the effectiveness and explainability.
    In this experiment, we adjust the granularity controlling hyper-parameter
    for segmentation model $\alpha$ as 0.2 to make the best fit for the dataset.} In this case,
    we use running and walking data with different conditional dimensions to train our model,
    while the remained hopping and jumping data are regarded as abnormal cases.

Fig~\ref{fig:mocap_reconstruction} shows the reconstruction results of walking, running, jumping, hopping with 
corresponding conditional data, and their corresponding heatmap.
In the running and walking series, the dashed line (reconstructed data) matches well with the solid line (origin data)
while reconstructions of jumping and hopping deviate far from the origin data. The results show
the ability of \method to reconstruct different categories of time series given corresponding
conditional dimensions.

\begin{figure*}[htbp]
    \centering
    \subfigure[]{
        \begin{minipage}{0.48\linewidth}
            \includegraphics[width=\textwidth, height=0.65\textwidth]{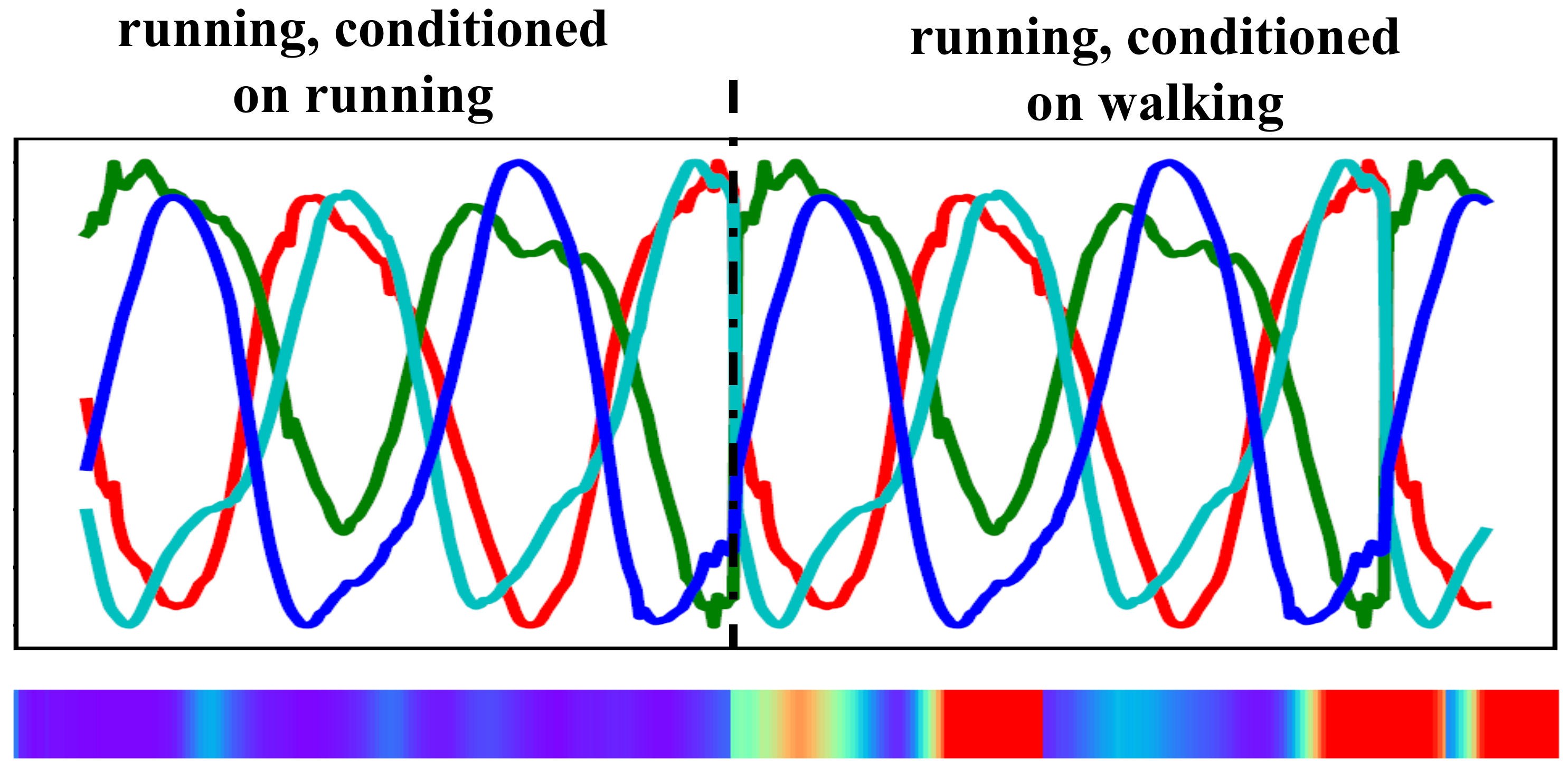}
        \end{minipage}
    }
    \hfill
    \subfigure[]{
        \begin{minipage}{0.48\linewidth}
            \includegraphics[width=\textwidth, height=0.65\textwidth]{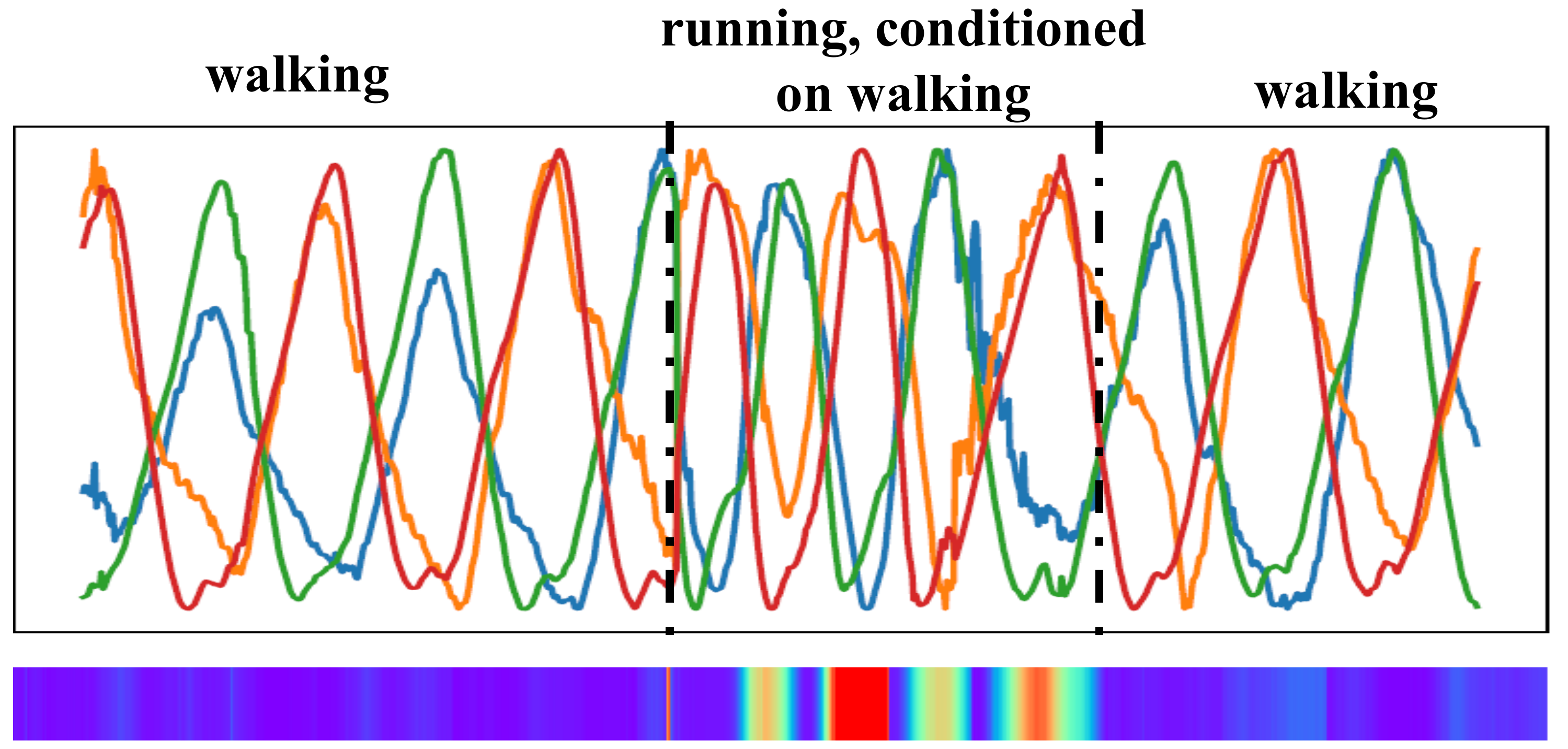}
        \end{minipage}
    }
    \label{fig:mocap_concat}
    \caption{Reconstruction of multi-category series with different conditional dimensions.}
\end{figure*}

Furthermore, to verify the effectiveness of conditional information, we concatenate two sequences.
The first one consists of two running cases labeled running and walking respectively.
The second one consists of two walking cases with one running case inserted to the middle whose conditional information is
labeled as walking. \atnn{The reconstruction error showed by heatmap (see Fig~\ref{fig:mocap_concat}) pinpoints
    both the mislabeled parts are not normal cases, which shows the effectiveness of conditional information.
    Heatmap points out the degree of deviation from the reconstructed line to the original line in detail,
    directing people’s attention straight  to the error district
    which reveals the explainability of our results.}
    
\begin{figure*}[htbp]
    \setlength{\belowcaptionskip}{-0.5cm}
    \subfigure[Distributions of anomalousness score.]{
        \begin{minipage}{0.49\linewidth}
            \includegraphics[width=\textwidth, height=0.7\textwidth]{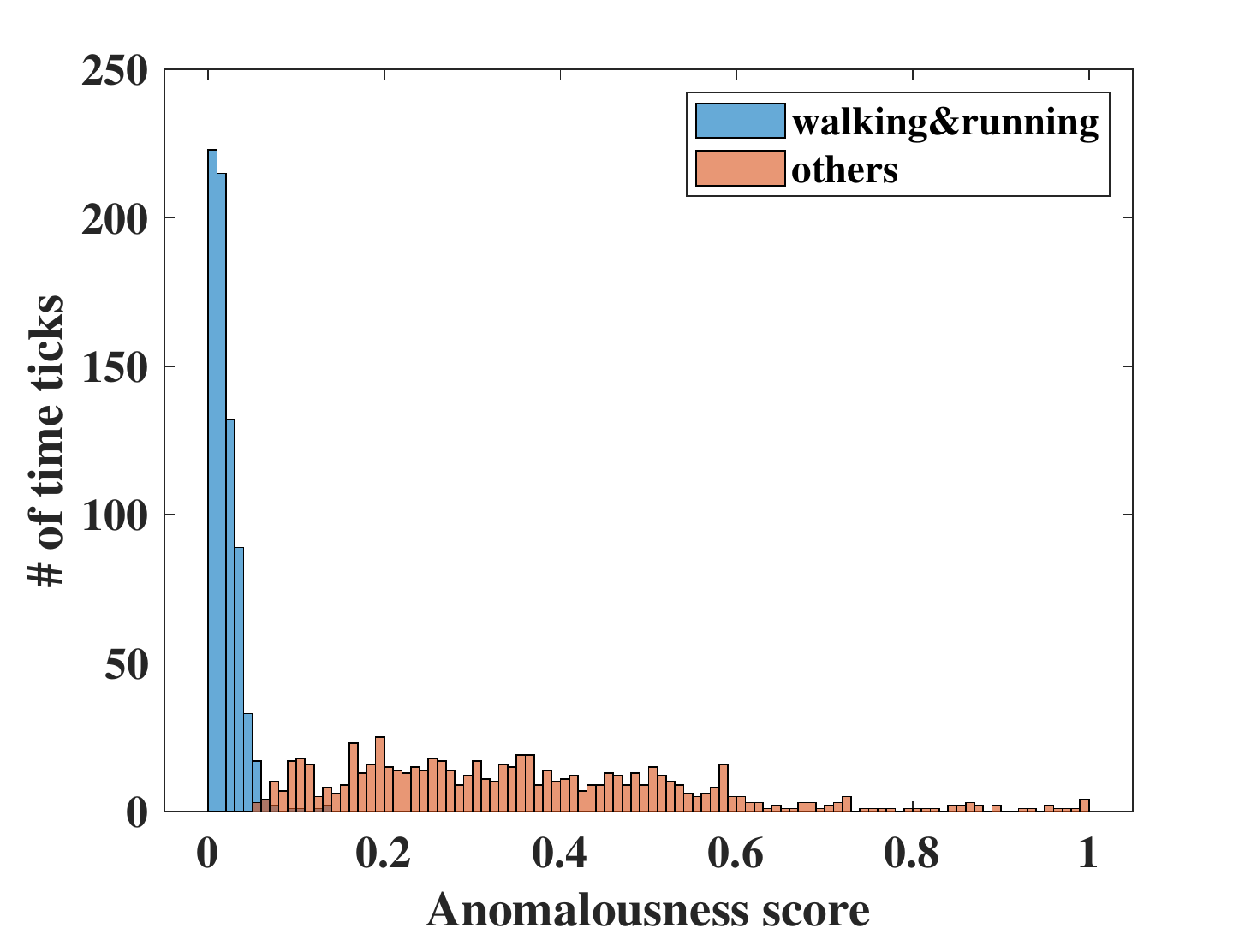}
        \end{minipage}
        \label{fig:mocap_distri}
    }
    \hfill
    \subfigure[Algorithm runs in linear time.]{
        \label{fig:time_consumption}
        \begin{minipage}{0.47\linewidth}
            \includegraphics[width=\linewidth, height=0.7\textwidth]{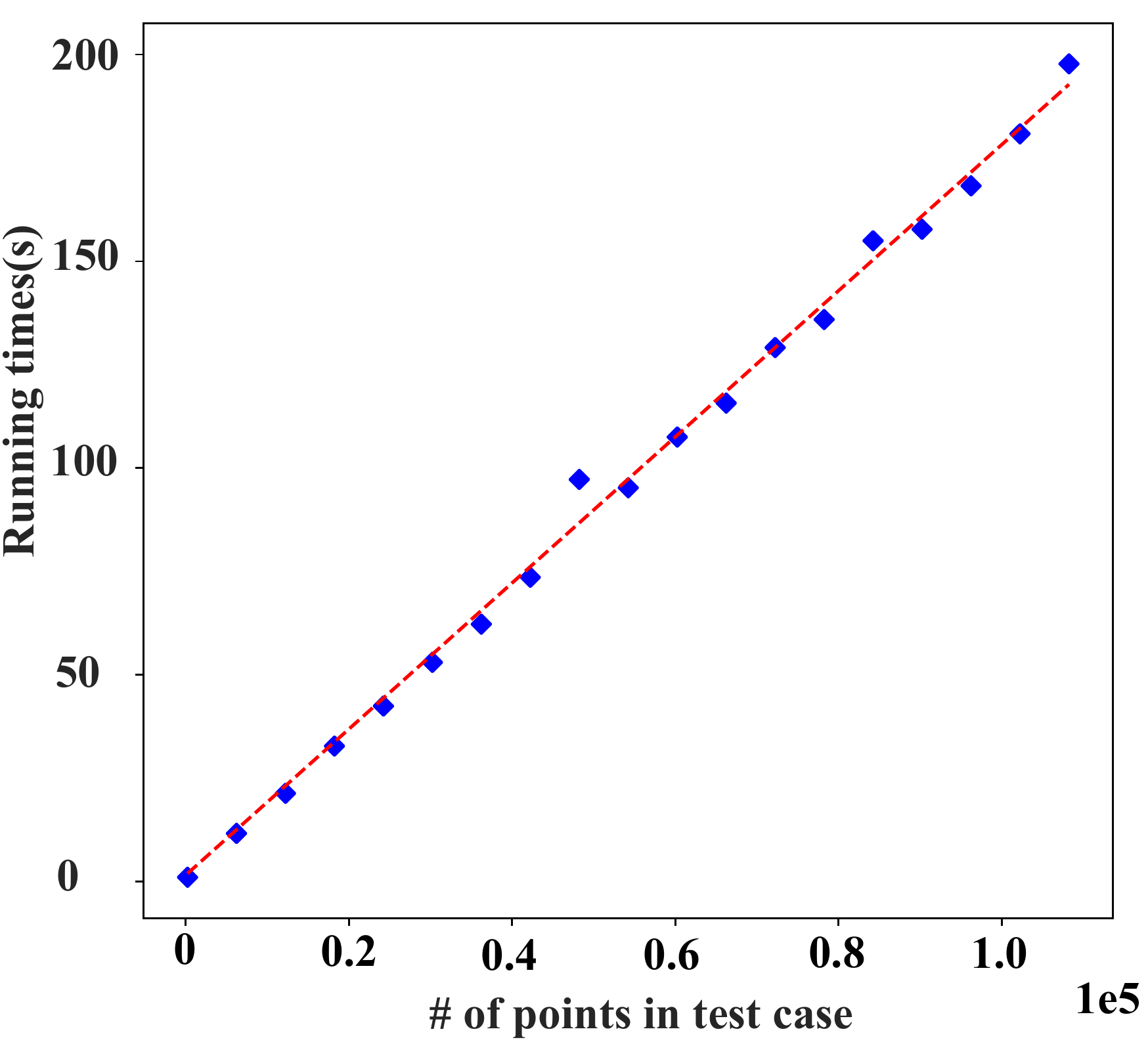}
        \end{minipage}
    }
    \caption{
        (a).  Distributions of anomalousness score
        between walking\&running and others.
        (b).Results display linear relation of running time
        and number of points which shows the scalability of \method.
    }
\end{figure*}
Histogram of anomalousness score is
shown in Fig~\ref{fig:mocap_distri}, which reveals the big differences between the distribution of scores of
walking\&running and others. Scores of walking\&running mainly gather below $0.1$ while the score of others
disperses widely from $0.1$ to $1$. These results show the great ability of \method in discriminating unusual
motions (jumping and hopping) by training with usual motions (walking and running).

\subsection{Convergence and Scalability (Q3)}

We run our trained model with a test set segmented into different lengths ranging from 240 to 120,000 and
record the timespan consumed during running. All the experiments are carried out on a server with a Tesla K80 GPU,
implemented in PyTorch. Results showed in Fig~\ref{fig:time_consumption} display the approximately linear relationship
between the number of points and running time which certifies the scalability of our \method.

\begin{figure*}[htbp]
    \centering
    \subfigure[Ideal F1 score - $\lambda$]{
        \label{fig:lambda_f1}
        \begin{minipage}{0.45\linewidth}
            \includegraphics[width=\linewidth, height=0.7\textwidth]{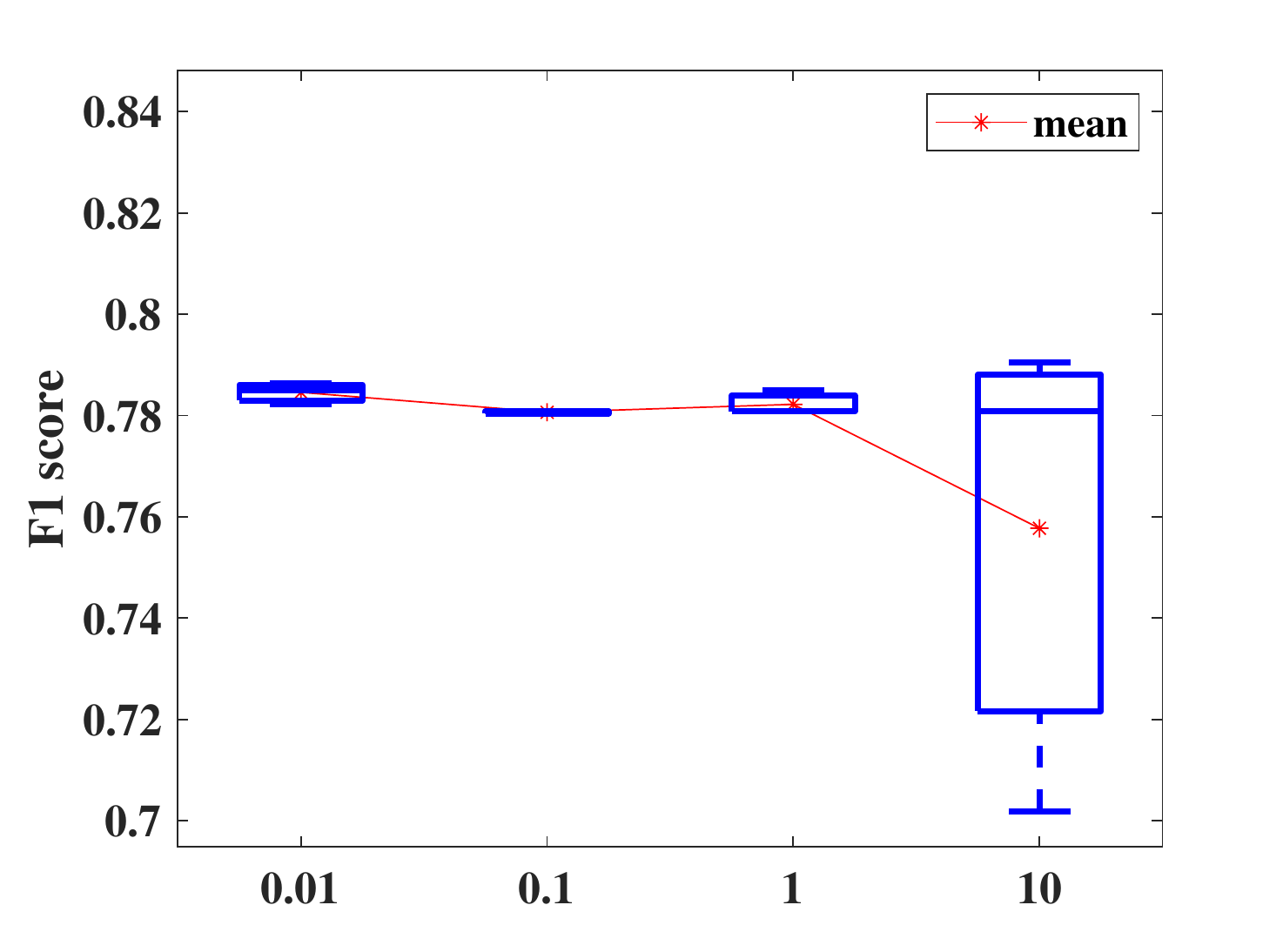}
        \end{minipage}
    }
    \hfill
    \subfigure[AUC score - $\lambda$]{
        \label{fig:lambda_auc}
        \begin{minipage}{0.45\linewidth}
            \includegraphics[width=\linewidth, height=0.7\textwidth]{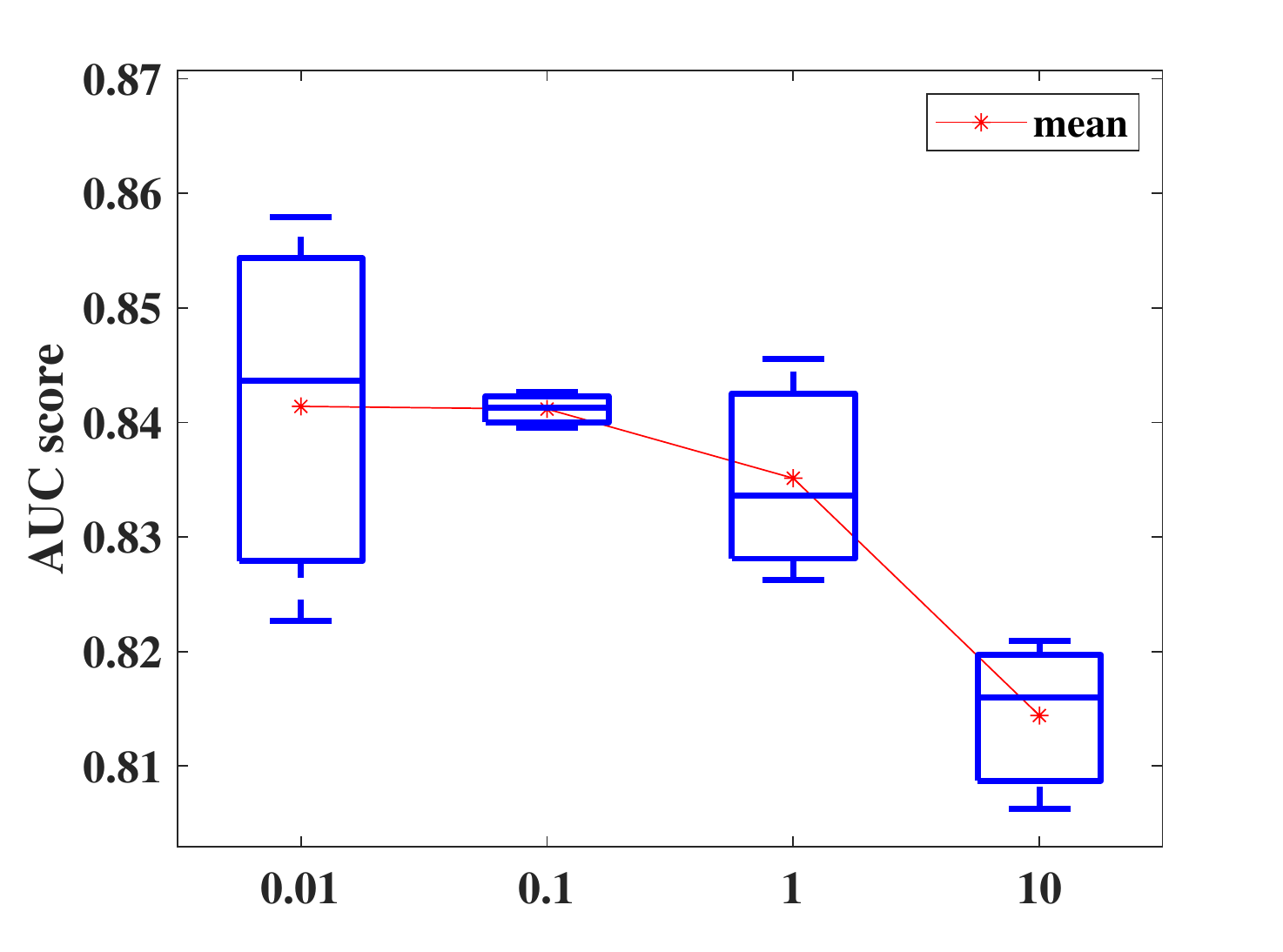}
        \end{minipage}
    }
    \caption{AUC and ideal F1 score of lambda experiments. Numbers on the x-axis stand for the lambda value.}
    \end{figure*}
\subsection{Parameter sensitivity (Q4)}

To ensure the best performance of our \method, we design architecture experiments concentrating on
\atn{the regularization parameter $\lambda$ and} dimensions reduced by PCA on SWaT dataset.

We evaluate the effect of regularization by assigning $\lambda$ the following values: 0.01, 0.1, 1, 10.
Fig~\ref{fig:lambda_f1} and Fig~\ref{fig:lambda_auc} depict the result of the regularization parameter experiment.
Although the highest ideal F1 score can be obtained at $\lambda = 0.01$, its severe
fluctuation cannot meet our request. On the contrary, results of $\lambda = 0.1$
have achieved both a relatively high F1 score and AUC score with a low fluctuation. Hence, we choose $0.1$ as our best $\lambda$.

\begin{figure*}
    \subfigure[Ideal F1 score - PCA.]{
        \label{fig:pca_f1}
        \begin{minipage}{0.45\linewidth}
            \includegraphics[width=\linewidth, height=0.7\textwidth]{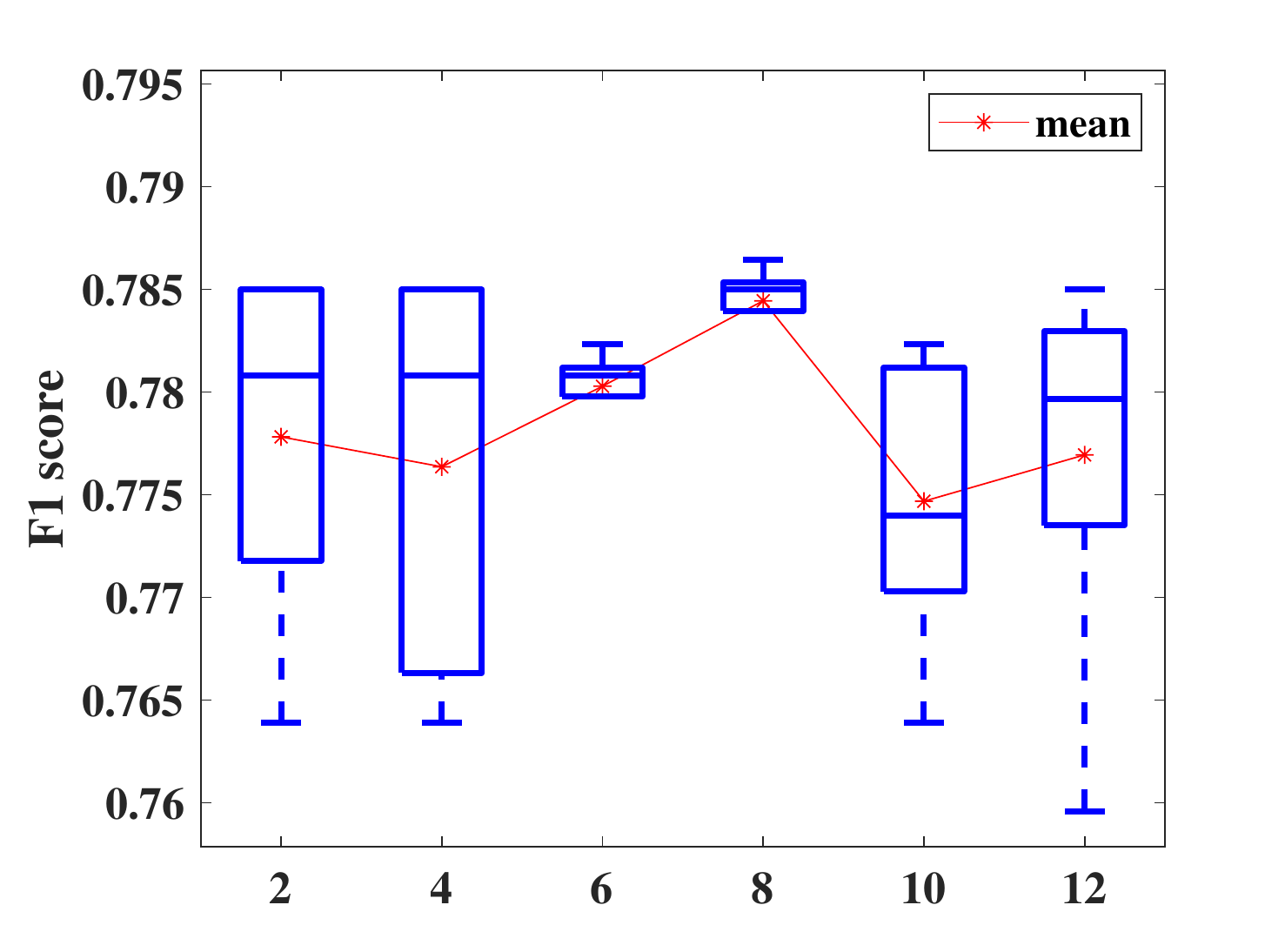}
        \end{minipage}
    }
    \hfill
    \subfigure[AUC score - PCA.]{
        \label{fig:pca_auc}
        \begin{minipage}{0.45\linewidth}
            \includegraphics[width=\linewidth, height=0.7\textwidth]{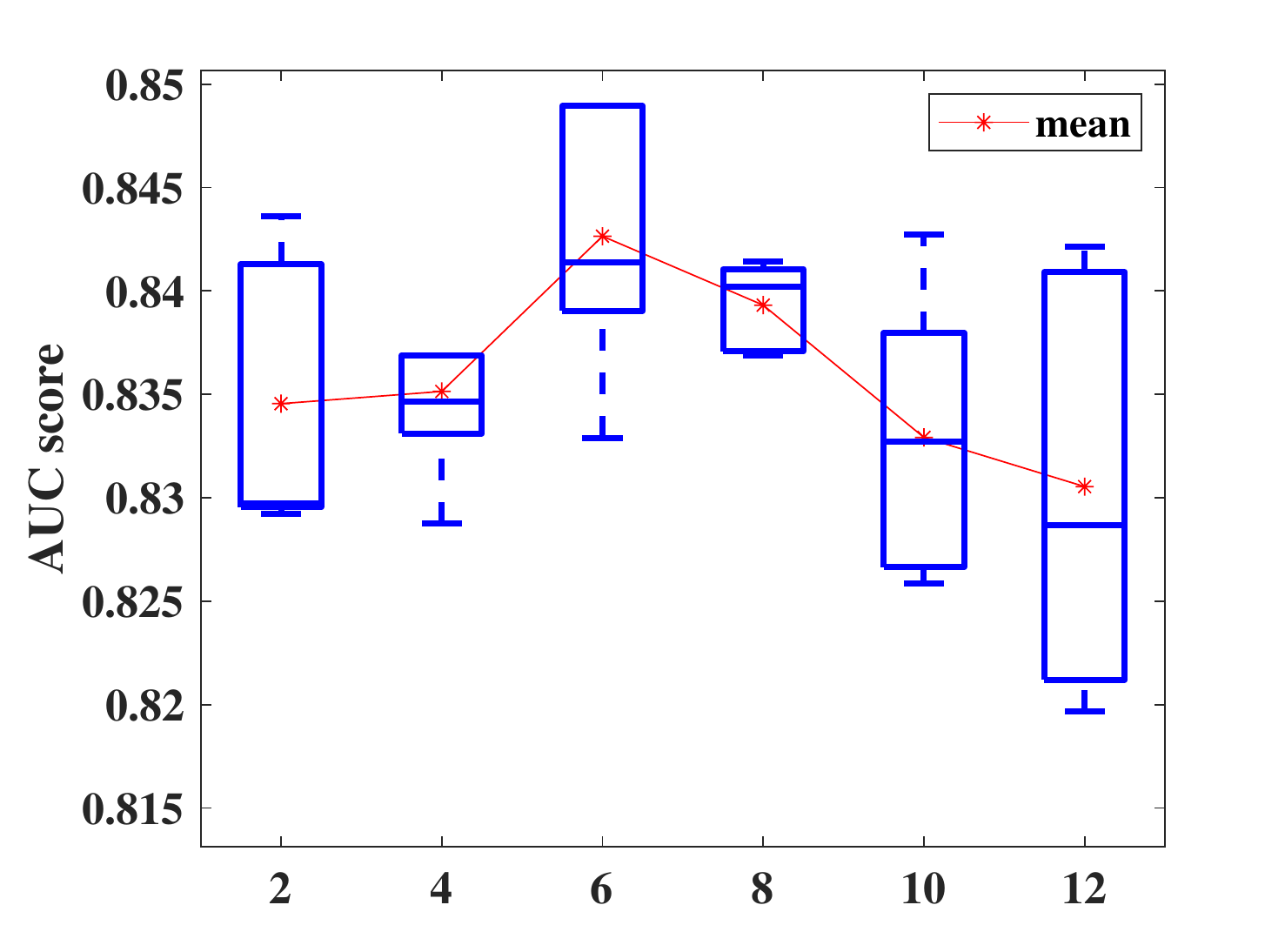}
        \end{minipage}
    }
    \caption{AUC and ideal F1 score of dimension reduction experiments.
        Model with PCA uses PCA to compress dimensions. Numbers on the x-axis stand for dimensions being reduced to.}
\end{figure*}

Candidates of reduced dimensions range from 2 to 12 with a stride of 2. Detailed results of box plots are shown in
\atnn{Fig~\ref{fig:pca_f1} and Fig~\ref{fig:pca_auc}.}
We can draw a conclusion that the dimension reduced to 6 by PCA has the best AUC score and 8 has
the best ideal F1 score. Due to ideal F1 score shall only appear based on well-adjusted parameters
and in considering of the generality, we choose dimension reduced to 6 by PCA as our best parameters.

\section{Conclusion}
We propose an anomaly detection algorithm for big time series based on reconstruction. 
Advantages of \method are as follows: 1) Multi-scale reconstruction: \method is trained from 
coarse to fine-grained segments for best reconstruction performance and \method is able 
to reconstruct multi-mode time series given different state conditions; 2) Effectiveness: 
\method outperforms baseline methods on ideal F1 score and AUC score with acceptable fluctuation; 
3) Explainability: \method pinpoint ticks of anomalies through displaying anomalousness 
score shown in Fig~\ref{fig:architecture}; 4) Scalability: \method runs linearly in the size of total time series.

To model flexible lengths of time series segments, we reconstruct them using GRU networks.
The inherent characteristics of GRU make it model well with 
smooth time series (i.e.,~spikes are abnormal).
However, this does not limit \method's applications since normally 
smooth time series occur in many domains such as infrastructure and traffic monitoring
and regularized motion analysis especially for the aged and mobility-impaired people.


\bibliography{acml21}




\end{document}